\documentclass[letterpaper, 10 pt, conference]{ieeeconf}
\IEEEoverridecommandlockouts                              

\usepackage{times}
\usepackage{epsfig}
\usepackage{graphicx}
\usepackage{amsmath}
\usepackage{amssymb}
\usepackage{graphicx}
\usepackage{amsmath}
\usepackage{booktabs}
\usepackage{algorithm}
\usepackage{algorithmic}
\usepackage{comment}
\usepackage{bm} 
\usepackage{multirow}

\usepackage[pagebackref=true,breaklinks=true,letterpaper=true,colorlinks,bookmarks=false]{hyperref}

\author{Jinchang Zhang, Arnold Zumbrun, Jing Lin, and Guoyu Lu
\thanks{Jinchang Zhang and Guoyu Lu are with the Intelligent Vision and Sensing (IVS) Lab at Indiana University Bloomington.
{\tt\small guoyulu62@gmail.com}. Arnold Zumbrun, Jing Lin, and Guoyu Lu are also with the US Air Force Research Lab (AFRL) during this work.}%
}
        
\begin{document}

\title{Foundation-Assisted Active Learning for Object Detection Annotation}

\maketitle
\begin{abstract}
The annotation cost for remote sensing object detection is high, while existing active learning methods still face several challenges in object detection scenarios, including the coupling of localization and classification uncertainty, severe localization noise in the cold-start stage, and pseudo-diversity caused by high-recall candidate proposals. To address these issues, we propose a foundation-model-collaborative active learning and semi-automatic annotation framework for efficient construction of remote sensing object detection datasets. We build a dual-source mechanism consisting of a reference localization source (SA-source) based on UPN+SAM2 and a detector prediction source (OD-source), and further propose a Foundation-model-enhanced Dual-Source Uncertainty estimation to improve sample selection quality in the cold-start stage by jointly modeling localization consistency and classification confidence. Furthermore, we propose Object-Centric Diversity Sampling, which constructs object-level representations using DINOv2 features and SAM2 masks to improve sample coverage while suppressing pseudo-diversity. To address geometric noise in the semi-automatic annotation stage, we design Dual-Source Box Switching, which replaces noisy detector boxes with matched refined boxes from the SA-source, thereby reducing the manual burden of box refinement. Experiments on DIOR, HRSC2016, DOTAv2, and FAIR1M show that our method achieves superior or comparable results under most annotation budgets, with notably stronger cold-start sample efficiency in the low-budget regime.
\end{abstract}

\section{Introduction}
Object detection heavily depends on large-scale, high-quality annotated data~\cite{lee2024coreset,wu2022entropy,fu2026dav}. In remote sensing, this process is often more time-consuming and costly due to factors such as dense small objects, large scale variations, and and complex imaging conditions~\cite{lin2025keypoint,lin20253d}. Active learning is widely regarded as an effective way to alleviate this challenge, as its core idea is to prioritize more informative samples for annotation under a limited labeling budget, thereby achieving better detection performance with lower human annotation cost, analogous to adaptive resource allocation in model optimization~\cite{tian2026curvatureadaptiveconsistencyflowmatching}.

\begin{figure*}[t]
\begin{center}
\includegraphics[width=17cm, height=8cm]{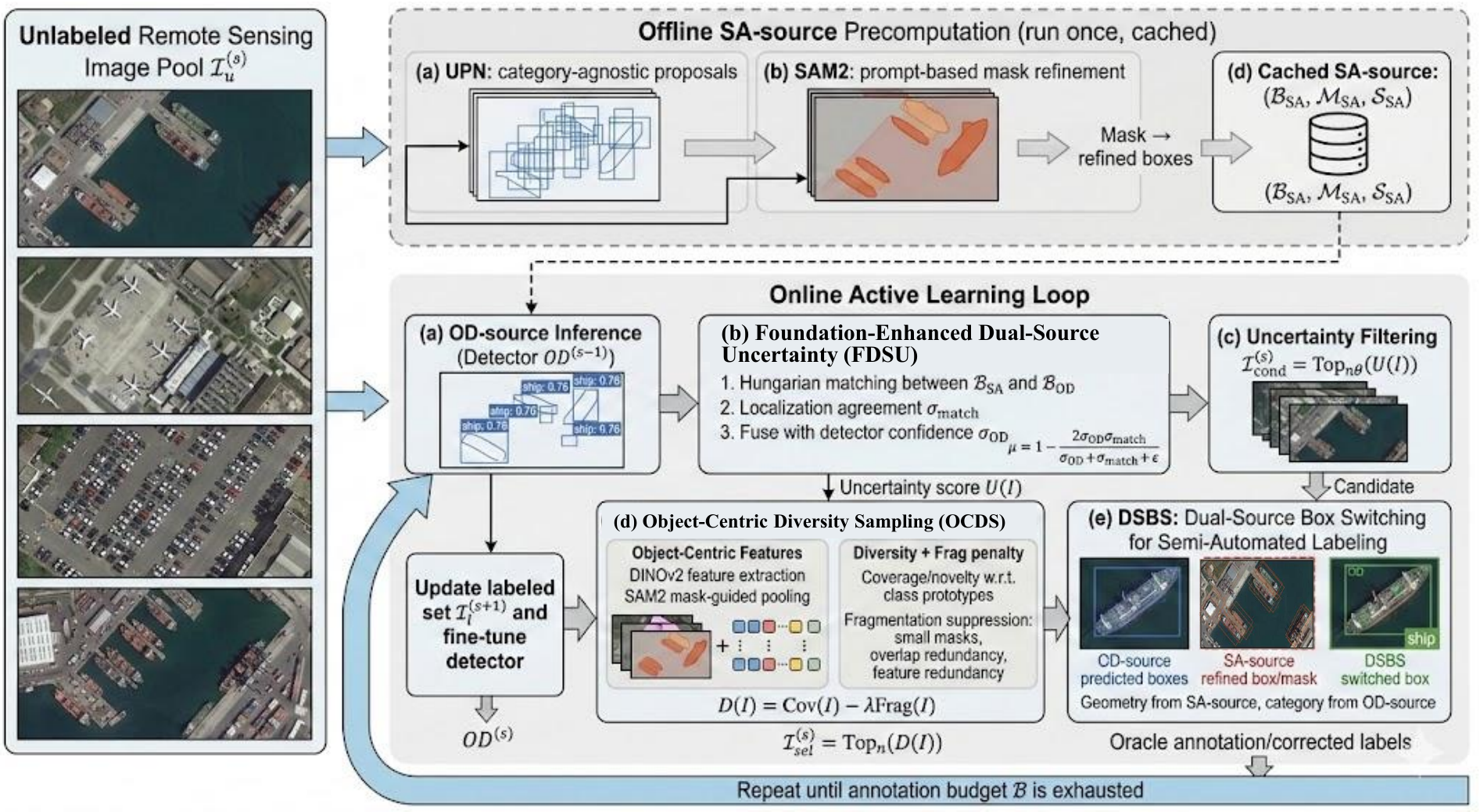}
\end{center}
\vspace{-6 mm}
\caption{Foundation-model-collaborative active learning and semi-automatic annotation framework for remote sensing object detection. The left panel shows the unlabeled remote sensing image pool. The upper branch performs offline SA-source precomputation (executed once and cached): UPN generates category-agnostic proposal boxes, and SAM2 refines them with masks (producing $\mathcal{B}_{SA}, \mathcal{M}_{SA}, \mathcal{S}_{SA}$). The lower branch is the online active learning loop: the detector runs inference on unlabeled images to produce the OD-source, followed by FDSU for uncertainty-based filtering; then, OCDS selects samples to annotate from the candidate set. During the semi-automatic annotation stage, DSBS replaces noisy OD-source boxes with matched refined boxes from the SA-source (geometry from the SA-source, category labels from the OD-source). Finally, the labeled set is updated and the detector is fine-tuned. This process repeats until the annotation budget is exhausted.}
\vspace{-7mm}
\label{overallframework}
\end{figure*}

Existing active learning methods mainly involve uncertainty estimation, diversity modeling, and data valuation~\cite{yoo2019learning,agarwal2020contextual,bar2024active,xiao2026points}. Although these methods have shown promising results, they still exhibit clear limitations in remote sensing object detection. First, in active learning for object detection, the value of a sample depends on both localization quality and category discrimination. However, existing methods often rely primarily on the detector’s own prediction confidence for sample selection, making it difficult to explicitly distinguish between classification uncertainty and localization uncertainty. This limitation is evident in the cold-start stage: benefiting from pretrained backbones, a detector’s classification capability may improve relatively quickly, while its localization remains noisy. As a result, sampling strategies based solely on detector confidence are easily affected by unstable bounding-box predictions and geometric bias~\cite{xiaopath}, reducing selection quality in the early stage of active learning. To alleviate this issue, we need a reference source that is relatively independent of detector predictions and provides more stable localization. Second, active learning must not only identify “hard samples,” but also ensure sample coverage under a limited annotation budget. However, a single model is often unable to simultaneously provide high-recall candidate generation, precise object localization, and robust semantic representations~\cite{zhang2025vision}. To address this, we introduce visual foundation models with complementary capabilities and build a collaborative pipeline spanning candidate generation, region refinement, and semantic representation, so that uncertainty estimation and diversity sampling in active learning can be grounded on higher-quality object representations rather than relying solely on the detector’s single prediction signal. Third, while the introduction of foundation models provides stronger candidate perception and representation capabilities for active learning, it also brings new forms of noise and redundancy. For example, model-generated candidates may contain over-fragmented instances, which can create “pseudo-diversity” during diversity sampling and reduce the effective use of the annotation budget. This implies that foundation models cannot simply be stacked onto an active learning pipeline; instead, they must be co-designed with the sampling strategy to suppress candidate redundancy and fragmentation noise. Finally, in the cold-start stage, even if category predictions are somewhat reliable, annotators must frequently correct boxes with large localization errors, leading to high practical interaction costs. Therefore, how to leverage foundation models to provide stable geometric priors and reduce human–computer interaction in semi-automatic annotation is key to deploying active learning in practice.

To address these challenges, we propose a foundation-model-collaborative framework (Fig \ref{overallframework}) for active learning and semi-automatic annotation to efficiently construct remote sensing datasets.  We build a dual-source mechanism: we use UPN+SAM2 to precompute and cache category-agnostic, high-recall, and localization-stable reference boxes (SA-source), while using current-round detector outputs as a learnable prediction source (OD-source). On top of this, Foundation-Enhanced Dual-Source Uncertainty (FDSU) matches bounding boxes from both sources and jointly models localization agreement and classification confidence to accurately assess sample informativeness. Furthermore, we perform mask-guided diversity sampling in the DINOv2~\cite{oquab2023dinov2} feature space and introduce a candidate fragmentation suppression term to avoid pseudo-diversity from redundant proposals. Finally, a Dual-Source Box Switching (DBS) mechanism replaces noisy detector boxes with SA-source refined boxes during the cold-start stage, reducing manual interaction costs and accelerating the active learning loop.

Overall, our contributions are summarized as follows:
1. We propose a foundation-model-collaborative active learning framework for remote sensing object detection, which integrates foundation-model capabilities into the active learning and semi-automatic annotation loop.
2. By matching bounding boxes between the two sources, we derive a sample uncertainty score that is better suited for active learning in object detection, improving sampling stability in the cold-start stage.
3. We design a redundancy penalty tailored to the introduction of visual foundation models, thereby improving annotation budget efficiency.
4. During the cold-start stage, we use SA-source refined boxes to replace detector-predicted boxes, which reduces annotation cost while improving detection performance.

\section{RELATED WORK}
\vspace{-1mm}
\subsection{Active Learning}
\vspace{-1mm}
Active learning for object detection mainly focuses on uncertainty and diversity modeling, but practical deployment must also address the cold-start stage, where scarce initial annotations cause unstable sample selection. Uncertainty-based methods select informative samples using prediction confidence—including loss prediction modules~\cite{yoo2019learning}, instance-/image-level uncertainty~\cite{yuan2021multiple}, and classification-localization inconsistency measures~\cite{jing2024object}. However, classification and localization errors are tightly coupled, making early-stage uncertainty estimation highly sensitive to localization noise. Diversity-based methods improve feature-space coverage via core-set selection~\cite{sener2017active}, clustering constraints~\cite{wang2017incorporating}, prototype-based representation alignment~\cite{xiao2026prototype}, relation-aware representation learning~\cite{li2024comae}, or hybrid dual-objectives like PPAL~\cite{yang2024plug}, USDM~\cite{yang2015multi}, and DivProto~\cite{wu2022entropy}. Yet, they still suffer from redundant candidates and ``pseudo-diversity" in high-recall proposal settings. In cold-start active learning, the detector lacks the labeled samples needed to form a stable data distribution representation, which biases uncertainty and distorts diversity estimation. To mitigate this, ALWOD~\cite{wang2023alwod} uses a generative warm-start with student-teacher disagreement, while Chen et al.~\cite{chen2022making} improve early-stage representativeness via self-supervised contrastive learning and K-means clustering. Overall, relying solely on detector predictions is insufficiently robust during cold-start, motivating the introduction of stronger priors or auxiliary mechanisms to improve sample selection.

\vspace{-1mm}
\subsection{Vision foundation models}
\vspace{-1mm}
Vision foundation models have reshaped downstream visual tasks. Large-scale pretraining and adaptation methods~\cite{caron2021emerging,oquab2023dinov2,simeoni2025dinov3,xiao2026reversible} endow vision transformers and vision-language models with powerful representation capabilities. Concurrently, task-oriented foundation models have emerged: SAM~\cite{kirillov2023segment,ravi2024sam} introduce a universal, prompt-based segmentation paradigm to produce class-agnostic masks, while the Universal Proposal Network (UPN)~\cite{jiang2024chatrex}, built on detection transformers~\cite{carion2020end,jiang2024t}, generates class-agnostic bounding boxes under varied objectness settings. Despite these individual advancements, active learning and vision foundation models \cite{zhang2025adaptive} lack deep collaborative design for remote sensing object detection. Existing methods like AL4FM~\cite{burges2025active} integrate segmentation models into active learning, but their foundation model utility remains limited. To address this, we integrate vision foundation model capabilities at three levels—candidate construction, sample selection, and semi-automatic annotation—to improve both cold-start sample efficiency and overall annotation efficiency in remote sensing object detection data construction.

\vspace{-1mm}
\section{Method} \label{sec:framework_overview}
\vspace{-1mm}
In this chapter, we introduce our foundation-model-driven active learning and semi-automatic annotation framework for accelerating remote sensing object detection dataset construction under a limited annotation budget. The core idea of the framework is to build a dual-source collaborative mechanism. On the one hand, we use UPN$\rightarrow$SAM2 to form a reusable, high-recall reference localization source (SA-source), which provides stable geometric priors during the cold-start stage. On the other hand, we use the current-round detector outputs as a learnable prediction source (OD-source), which provides category predictions and confidence information. Based on these two sources, we adopt a two-stage sample selection strategy, namely uncertainty filtering + diversity sampling, to select the most valuable samples for annotation. Moreover, we introduce a dynamic box-switching mechanism to reduce manual box-correction cost in early-stage annotation, thereby improving the efficiency and stability of the active learning loop.
\subsection{SA-Source Precomputation} \label{sec:sa_source_precompute}
This section introduces the SA-source in our framework: a reusable, category-agnostic, high-recall localization candidate pool that provides stable spatial priors for subsequent Foundation-Enhanced Dual-Source Uncertainty and semi-automatic annotation. Unlike approaches that rely on early detector predictions, the SA-source is constructed by a two-stage foundation-model pipeline. We first use the Universal Proposal Network (UPN) to generate category-agnostic proposal boxes, and then use SAM2 to refine these proposals into masks, from which tight bounding boxes are derived. This process is executed offline only once per image, and the cached results are reused across all active learning rounds.

\subsubsection{Category-Agnostic Proposal Generation (UPN Proposals)}
\label{sec:upn_proposals}

Given an unlabeled image $I$, UPN outputs a set of category-agnostic proposal boxes with objectness scores:
$
\mathrm{UPN}(I)$ $\rightarrow $$P(I)=\left\{ \left(b_j^{u}, s_j^{\mathrm{upn}}\right) \right\}_{j=1}^{N_u(I)},
$
where $b_j^{u}=(x_1^j, y_1^j, x_2^j, y_2^j)$ denotes the coordinates of the $j$-th proposal box, $s_j^{\mathrm{upn}}$ $\in$ $(0,1)$ is the category-agnostic objectness confidence, and $N_u(I)$ is the number of proposals generated by UPN. The role of UPN is to achieve high recall over potential objects, thereby avoiding sole reliance on unstable localization predictions during the detector cold-start stage.
To control computational cost and reduce obviously noisy candidates, we apply a lightweight pre-filtering step to $P(I)$ and retain the Top-$K_u$ proposals:
$
P_{K_u}(I) = \mathrm{Top}\text{-}K_u\!\left(\left\{s_j^{\mathrm{upn}}\right\}\right).$
Optional geometric constraints can also be applied, such as an area threshold and an aspect-ratio range:
$
{\mathrm{area}(b_j^{u})}/{HW} \ge \tau_b,
\rho_{\min} \le {w(b_j^{u})}/{h(b_j^{u})} \le \rho_{\max}.
$
These filters are only used to remove extreme outliers. The core fragmentation suppression is explicitly modeled later in the diversity stage (Sec.~\ref{sec:mask_guided_diversity}).

\subsubsection{Mask Refinement and Tight Box Extraction (SAM2 Refinement)}
\label{sec:sam2_refinement}

For each retained proposal box $b_j^{u} \in P_{K_u}(I)$, we use it as a prompt to SAM2 and generate an object mask:
$
\mathrm{SAM2}(I, b_j^{u}) \rightarrow m_j,
$
where $m_j \in \{0,1\}^{H \times W}$ is a binary mask. We then derive a tight bounding box from the mask:
$
b_j^{sa} = \mathrm{bbox}(m_j),
$
where $\mathrm{bbox}(\cdot)$ denotes the minimum enclosing rectangle of the foreground pixel set in the mask. This ``box $\rightarrow$ mask $\rightarrow$ tight box'' procedure significantly improves the alignment between proposals and true object boundaries, providing a more reliable reference for localization-consistency measurement in subsequent FDSU.

\subsubsection{SA-Source Caching and Reuse}
\label{sec:sa_cache_reuse}

In summary, for image $I$, we obtain and cache the following SA-source triplets:
$
B_{SA}(I)$=$\left\{b_j^{sa}\right\}_{j=1}^{N_{SA}(I)}$, 
$M_{SA}(I)$=$\left\{m_j\right\}_{j=1}^{N_{SA}(I)}$, 
$S_{SA}(I)$=$\left\{s_j^{\mathrm{upn}}\right\}_{j=1}^{N_{SA}(I)},
$
where $N_{SA}(I) \le K_u$.
The cache is generated only once per image and is directly loaded in each active learning round. As a result, the foundation-model inference cost is reduced from ``repeated computation every round'' to a ``one-time offline preprocessing'' cost, improving the scalability of the overall framework for large-scale remote-sensing data annotation scenarios.

\vspace{-1mm}
\subsection{Foundation-Enhanced Dual-Source Uncertainty
} \label{sec:dsue}
\vspace{-1mm}
In active learning for object detection, the ``informativeness'' of a sample comes not only from uncertainty in category discrimination, but also from uncertainty in localization quality. This is especially important in the cold-start stage: the detector's classification capability may improve quickly due to pretraining, while localization remains relatively unstable. Therefore, using detector confidence alone as the sampling criterion often overlooks localization noise, which is a key factor in this stage. To address this issue, we propose Foundation-Enhanced Dual-Source Uncertainty (FDSU). FDSU uses the SA-source (reference boxes precomputed by UPN$\rightarrow$SAM2) to provide a stable localization reference, and uses the OD-source (current-round detector predictions and classification confidence) to reflect the model's current belief. We then fuse localization agreement and classification confidence into a unified uncertainty measure, which is used to filter a candidate set $\mathcal{I}^{(s)}_{\mathrm{cond}}$.

\subsubsection{Definition of the SA/OD Dual Sources}
\label{sec:dsue_sources}

For any unlabeled image $I \in \mathcal{I}^{(s)}_u$, the cached SA-source is
$
\mathcal{B}_{SA}(I)=\{b^{sa}_j\}_{j=1}^{N_{SA}(I)},
$
where $b^{sa}_j$ is the tight box derived from the SAM2 mask (Sec.~\ref{sec:sa_source_precompute}).
The OD-source predictions from the current-round detector $\mathrm{OD}^{(s-1)}$ are
$
\mathrm{OD}^{(s-1)}(I)\rightarrow
\mathcal{B}_{OD}(I)=\{b^{od}_k\}_{k=1}^{N_{OD}(I)},
\Sigma_{OD}(I)=\{\sigma^{od}_k\},
\label{od}
$
where $\sigma^{od}_k\in(0,1)$ is the classification confidence for box $b^{od}_k$ (e.g., maximum class probability or objectness $\times$ class probability).

\subsubsection{Hungarian Matching and Localization Agreement}
\label{sec:dsue_matching}

Since $\mathcal{B}_{SA}(I)$ and $\mathcal{B}_{OD}(I)$ generally differ in both cardinality and ordering, we align the two sets using one-to-one Hungarian matching while ignoring category labels, so as to focus solely on localization agreement.
We first construct a cost matrix $\mathbf{C}\in\mathbb{R}^{N_{SA}\times N_{OD}}$ using an IoU-induced cost:
\vspace{-1mm}
\begin{equation}
\mathbf{C}_{jk}=1-\mathrm{IoU}(b^{sa}_j,\; b^{od}_k),
\qquad
\mathrm{IoU}(a,b)=\frac{|a\cap b|}{|a\cup b|}.
\vspace{-2mm}
\label{eq:dsue_cost}
\end{equation}

The Hungarian algorithm then produces a matching set with minimum total cost,
$
\mathcal{P}(I)=\{(j,k)\}.
$
For unmatched boxes (e.g., extra candidates on one side), we do not include them in FDSU fusion; however, they can still be used in the diversity stage (see Sec.~\ref{sec:mask_guided_diversity}).
Given a matched pair $(j,k)\in\mathcal{P}(I)$, we convert the matching cost into a localization agreement quality score $\sigma^{\mathrm{match}}_{jk}\in(0,1]$. 

\subsubsection{Fusing Classification Confidence and Localization Agreement (Object-level Uncertainty)}
\label{sec:dsue_object_level}

Localization agreement alone is still insufficient to characterize {learnable informativeness}: some objects may have stable localization but remain difficult to classify, and are thus still valuable for annotation. We therefore fuse the localization agreement $\sigma^{\mathrm{match}}_{jk}$ and the OD-source classification confidence $\sigma^{od}_k$ to obtain an object-level uncertainty $\mu_{jk}$.
We use the harmonic mean to emphasize the ``short-board effect'' (i.e., a low value in either term leads to higher uncertainty):
\vspace{-1mm}
\begin{equation}
\mu_{jk}
=
1-\frac{2\,\sigma^{od}_k\,\sigma^{\mathrm{match}}_{jk}}
{\sigma^{od}_k+\sigma^{\mathrm{match}}_{jk}+\epsilon},
\qquad
\epsilon=10^{-6}.
\vspace{-1mm}
\label{eq:dsue_object_uncertainty}
\end{equation}


\subsubsection{Image-level Uncertainty and Candidate Set Filtering}
\label{sec:dsue_image_level}

To select a candidate set for subsequent diversity sampling from the unlabeled pool, we aggregate object-level uncertainty into an image-level uncertainty score $U(I)$. Since remote sensing images often contain multiple instances with large scale variations, simple averaging may be diluted by many easy objects. To improve robustness, we adopt Top-$K$ aggregation (where $K$ can be a constant, e.g., 10, or adaptively chosen based on the number of instances):
$
U(I)=\frac{1}{K}\sum_{(j,k)\in \mathrm{TopK}(\mathcal{P}(I))}\mu_{jk},
\label{eq:dsue_image_uncertainty}
$
where $\mathrm{TopK}(\mathcal{P}(I))$ denotes the top $K$ matched pairs in $\mathcal{P}(I)$ ranked by $\mu_{jk}$ in descending order. If the number of matched pairs is smaller than $K$, we average over all matched pairs.
Given the per-round annotation budget $n$ and a candidate expansion factor $\theta>1$, we select the top $n\theta$ most uncertain images from the unlabeled set to form the candidate set:
$
\mathcal{I}^{(s)}_{\mathrm{cond}}
=
\operatorname{Top}_{n\theta}\Big(\{U(I)\}_{I\in\mathcal{I}^{(s)}_u}\Big).
\label{eq:dsue_candidate_set}
$
We then perform diversity sampling on $\mathcal{I}^{(s)}_{\mathrm{cond}}$ (Sec.~\ref{sec:mask_guided_diversity}) to obtain the final set for annotation, $\mathcal{I}^{(s)}_{\mathrm{sel}}$.
For each image, the Hungarian matching complexity is
$
O\!\big(\min(N_{SA},N_{OD})^3\big).
$
In practice, the number of SA-source candidates can be controlled by Top-$K_u$ (Sec.~3.1), and the number of OD-source prediction boxes can also be limited by keeping Top-$K_{od}$ after NMS. Therefore, FDSU can be implemented efficiently in practice.
More importantly, the SA-source is obtained from offline caches and does not require rerunning UPN and SAM2 in each AL round. As a result, the computational cost is concentrated on detector inference and matching-based fusion, making the framework practical for large-scale remote sensing annotation.

\vspace{-2mm}
\subsection{Object-Centric Diversity Sampling (OCDS)}
\label{sec:mask_guided_diversity}
\vspace{-1mm}
In Sec.~\ref{sec:dsue}, we use FDSU to filter a high-uncertainty candidate set $\mathcal{I}^{(s)}_{\mathrm{cond}}$ from the unlabeled pool. However, uncertainty-prioritized sampling tends to produce ``near-duplicate'' samples (images from the same geographic region, under the same sensor condition, or with highly similar target appearance), which reduces the coverage efficiency of the annotation budget. Therefore, in this section, we further perform diversity sampling on $\mathcal{I}^{(s)}_{\mathrm{cond}}$ to maximize coverage over object appearance and scene conditions.
Unlike methods that rely only on SAM masks, we use SAM2 masks to suppress background interference, and build object-level representations and class prototypes in the DINOv2 semantic space for more stable coverage estimation. Furthermore, considering that UPN proposals may exhibit {over-fragmentation} (many local small boxes or duplicate boxes), we introduce an explicit fragmentation suppression term to prevent diversity scores from being dominated by noisy candidates.

\subsubsection{Candidate Objects and Object Embeddings}
\label{sec:div_candidate_embeddings}

For any candidate image $I \in \mathcal{I}^{(s)}_{\mathrm{cond}}$, we obtain candidate masks and bounding boxes from the cached SA-source:
$
\mathcal{M}_{SA}(I)=\{m_j\}_{j=1}^{N_{SA}(I)},
\mathcal{B}_{SA}(I)=\{b^{sa}_j\}_{j=1}^{N_{SA}(I)}.
$
To reduce computational cost and avoid extreme noisy candidates, we retain only the Top-$K$ candidate objects (ranked by $s^{\mathrm{upn}}$, mask area, or simple geometric filtering), denoted by $\mathcal{O}(I)$. That is,
$
\mathcal{O}(I)=\{(b_j,m_j)\}_{j=1}^{N_I},
N_I\le K.
$
We then extract object-level embeddings. Let the DINOv2 feature map for the full image be
$
F(I)\in \mathbb{R}^{C\times H'\times W'}.
$
We downsample the mask $m_j$ to the feature-map resolution, denoted by $m_j^{\mathrm{down}}$, and apply mask-guided pooling to obtain the object feature:
\vspace{-1mm}
\begin{equation}
\footnotesize
f_j(I)=
\frac{\sum_{u,v} F(I)[:,u,v]\cdot m_j^{\mathrm{down}}[u,v]}
{\sum_{u,v} m_j^{\mathrm{down}}[u,v]+\varepsilon},
\bar f_j(I)=\frac{f_j(I)}{\|f_j(I)\|_2}.
\vspace{-2mm}
\label{eq:div_mask_pooling}
\end{equation}


\subsubsection{Class Prototype Construction}
\label{sec:div_class_prototypes}

To measure how well candidate objects are covered by the {labeled distribution}, we construct class prototypes on the labeled set $\mathcal{I}^{(s)}_l$. Specifically, for each ground-truth box in each labeled image, we also use SAM2 to obtain a precise mask and compute an object embedding according to Eq.~\eqref{eq:div_mask_pooling}. For each class $c\in\{1,\dots,C\}$, we collect the embedding set:
$
\mathcal{F}^{(s)}_c
=
\{\bar f \mid \bar f \text{ comes from } \mathcal{I}^{(s)}_l \text{ and has class } c\}.
$
To capture intra-class diversity, we use k-means++ (with cosine distance) to cluster $\mathcal{F}^{(s)}_c$ into $p$ prototypes:
$
\mathcal{P}^{(s)}_c=\{p_{c,t}\}_{t=1}^{p},
\|p_{c,t}\|_2=1.
$
Here, $p$ is the number of prototypes per class (e.g., $p=5$). These prototypes represent ``already covered appearances'' and are used to quantify the coverage gain of new candidate objects.
The class label of each candidate object is obtained from the OD-source prediction. For candidate object $j$, we denote its predicted class by $\hat c_j$ (e.g., assigned by the detector box with maximum IoU with $b^{sa}_j$, or directly by detector predictions on the corresponding region).

\subsubsection{Coverage / Novelty Scoring}
\label{sec:div_novelty_scoring}

We define the degree to which a candidate object is {not yet covered} by prototypes of its predicted class as {novelty}:
$
\mathrm{novel}(j)=
\min_{t\in\{1,\dots,p\}}
\Big(1-\cos(\bar f_j(I),\, p_{\hat c_j,t})\Big),$
where $\cos(\cdot,\cdot)$ denotes cosine similarity. If $\bar f_j$ is  close to some prototype, then that appearance has already been covered in the labeled set and $\mathrm{novel}(j)$ is small; otherwise, it is large, indicating a potentially new appearance or scene with higher annotation value.
We aggregate object-level novelty into an image-level coverage score. To avoid mean dilution in dense-object images, we adopt Top-$K_d$ aggregation:
$
\mathrm{Cov}(I)
=
\frac{1}{K_d}
\sum_{j\in \mathrm{Top}K_d(\mathcal{O}(I))}
\mathrm{novel}(j),
\label{eq:div_coverage}
$
where $\mathrm{Top}K_d(\mathcal{O}(I))$ denotes the top $K_d$ objects ranked by $\mathrm{novel}(j)$ in descending order.

\subsubsection{Explicit Fragmentation Suppression}
\label{sec:div_fragmentation}

The high-recall property of UPN may introduce over-fragmented candidates (many local small boxes or highly overlapping duplicates). These candidates are often similar to each other in feature space and provide limited annotation gain, but can artificially inflate $\mathrm{Cov}(I)$ and create ``pseudo-diversity.'' To address this issue, we define a fragmentation penalty $\mathrm{Frag}(I)$ for each image to suppress candidate redundancy.

\textbf{(a) Small-area fragment ratio.}
Let $|m_j|$ denote the mask area (in pixels). We define
\vspace{-2mm}
\begin{equation}
r_{\text{small}}(I)
=
\frac{1}{|\mathcal{O}(I)|}
\sum_{j\in\mathcal{O}(I)}
\mathbb{1}\!\left(\frac{|m_j|}{HW}<\tau_a\right),
\vspace{-2mm}
\label{eq:div_rsmall}
\end{equation}
where $\tau_a$ is the small-area threshold.

\textbf{(b) Spatial redundancy (overlapping duplicates).}
\vspace{-2mm}
\begin{equation}
r_{\text{iou}}(I)
=
\frac{1}{|\mathcal{O}(I)|}
\sum_{j\in\mathcal{O}(I)}
\max_{k\neq j}\mathrm{IoU}(b_j,b_k).
\vspace{-2mm}
\label{eq:div_riou}
\end{equation}

\textbf{(c) Semantic redundancy (feature duplication).}
\vspace{-2mm}
\begin{equation}
r_{\text{sim}}(I)
=
\frac{1}{|\mathcal{O}(I)|}
\sum_{j\in\mathcal{O}(I)}
\max_{k\neq j}\cos(\bar f_j(I),\bar f_k(I)).
\vspace{-2mm}
\label{eq:div_rsim}
\end{equation}
\vspace{-2mm}
\textbf{(d) Candidate count penalty.}
\begin{equation}
r_{\text{cnt}}(I)=\log\big(1+|\mathcal{O}(I)|\big).
\vspace{-2mm}
\label{eq:div_rcnt}
\end{equation}

\subsubsection{Final Diversity Score and Selection}
\label{sec:div_final_selection}

Finally, we combine the coverage score and fragmentation penalty to define the image-level diversity score:
$
D(I)=\mathrm{Cov}(I)-\lambda\,\mathrm{Frag}(I),
$
where $\lambda\ge 0$ controls the strength of fragmentation suppression. Intuitively, $D(I)$ encourages selecting images that contain uncovered appearances/scenes while avoiding candidate redundancy caused by UPN over-fragmentation.
On the candidate set $\mathcal{I}^{(s)}_{\mathrm{cond}}$, we select the top $n$ images with the highest diversity scores as the final annotation set for the current round:
$
\mathcal{I}^{(s)}_{\mathrm{sel}}
=
\operatorname{Top}_{n}\Big(\{D(I)\}_{I\in\mathcal{I}^{(s)}_{\mathrm{cond}}}\Big).
$
We typically limit the number of candidate objects per image to $N_I\le K$ (e.g., $K\in[30,100]$), and when computing $r_{\text{iou}}$ and $r_{\text{sim}}$, we only consider a local neighbor set for each object (e.g., Top-$M$ overlapping boxes or Top-$M$ nearest feature neighbors), so that the overall complexity remains tractable.

\subsection{Dual-Source Box Switching (DSBS)}
\label{sec:dbs}

In the early iterations of active learning, even when the detector uses a pretrained backbone, its classification ability often improves relatively quickly, while localization quality remains significantly insufficient. This cold-start phenomenon---``classification converges earlier, localization converges later''---directly increases annotation cost: if the annotation interface displays detector-predicted boxes by default, annotators must frequently drag, resize, or redraw bounding boxes to correct localization errors, which offsets the efficiency gains brought by active learning. To reduce the manual box-correction burden in early rounds, we propose a {Dual-Source Box Switching} (DSBS) module. During annotation, DSBS replaces noisy detector boxes with more stable SA-source geometric boxes while preserving detector-provided category supervision, thus enabling semi-automated labeling with ``geometry from SA, category from OD.''
For any image to be annotated, $I \in \mathcal{I}^{(s)}_{\mathrm{sel}}$, we have:SA-source $
\mathcal{B}_{SA}(I)=\{b_j^{sa}\}_{j=1}^{N_{SA}(I)},
$,
OD-source (detector) \ref{od}.
The goal of DSBS is to generate, for each displayed object in the annotation interface, a {pre-annotation} box-label pair $(\tilde b, \tilde y)$ such that:
 $\tilde b$ is as close as possible to the true object boundary (to reduce geometric correction cost).
 $\tilde y$ remains consistent with the detector output (to preserve  supervision signals).

\subsubsection{Box Switching: Aligning OD Boxes to SA Boxes}
\label{sec:dbs_switching}

We first establish correspondences between $\mathcal{B}_{OD}(I)$ and $\mathcal{B}_{SA}(I)$. To ensure one-to-one assignment and avoid many-to-one conflicts, we adopt the same Hungarian matching scheme \ref{eq:dsue_cost}. 
This yields a matching set
$
\mathcal{P}(I)=\{(k,j)\}.
$
For each matched pair $(k,j)$, the switched pre-annotation box as:
\vspace{-1mm}
\begin{equation}
\tilde b_k=
\begin{cases}
b_j^{sa}, & \text{if } \mathrm{IoU}(b_k^{od}, b_j^{sa}) \ge \tau_{sw},\\[4pt]
b_k^{od}, & \text{otherwise},
\end{cases}
\vspace{-2mm}
\label{eq:dbs_switch}
\end{equation}
where $\tau_{sw}$ is the switching threshold used to avoid incorrectly replacing a detector box with an obviously unrelated SA box. Intuitively, when an OD box sufficiently overlaps with an SA box, we regard them as referring to the same object instance, and replacing the OD box with the SA box can significantly improve geometric quality.
The category label and confidence remain from the OD-source:
$
\tilde y_k=\hat y_k,
\tilde \sigma_k=\sigma_k^{od}.
$
Therefore, each detected instance is presented in the annotation interface as $(\tilde b_k,\tilde y_k)$.

\subsubsection{Unmatched Cases and High-Confidence Pre-Labeling}
\label{sec:dbs_unmatched}

The detector may produce unmatched prediction boxes (e.g., SA-source does not cover them, or related SA candidates were filtered), and the SA-source may also contain unmatched high-quality candidates (e.g., detector misses). We handle them separately:
\textbf{Unmatched OD boxes.}
We keep $(b_k^{od}, \hat y_k)$ unchanged for display, to avoid missing targets already discovered by the detector.
\textbf{Unmatched SA boxes.}
They can be optionally displayed as ``objects to confirm,'' which is especially useful in remote sensing scenes with dense small objects for recovering missed detections. To control annotation burden, we only display candidates with high objectness and simple geometric filtering:
\vspace{-0mm}
\begin{equation}
\left\{
b_j^{sa}\;\middle|\;
j\notin \pi_2(\mathcal{P}(I)),
\;
s_j^{\mathrm{upn}}\ge \tau_{\mathrm{upn}},
\;
\mathrm{area}(b_j^{sa})\ge \tau_b
\right\},
\vspace{-1mm}
\label{eq:dbs_unmatched_sa}
\end{equation}
where $\pi_2(\mathcal{P}(I))$ denotes the set of SA indices that appear in the matched pairs. For such candidates, the category can be initialized as ``unknown,'' or optionally assigned a weak label using the detector's maximum class response in that region.
In addition, DSBS can be combined with a confidence threshold for semi-automatic pre-labeling: for high-confidence instances with $\tilde \sigma_k \ge \tau_{cls}$, we directly provide pre-labels and the annotator only needs confirmation or minor edits; for low-confidence instances, the interface can prompt closer category inspection.

\subsubsection{Dynamic Scheduling: When to Enable/Disable DSBS}
\label{sec:dbs_schedule}

The benefit of DSBS is most significant in the cold-start stage. To make the switching {dynamic}, we provide two equivalent scheduling strategies:
{Fixed-round strategy.}
Enable DSBS for the first $T$ rounds and disable it afterward:
$
\text{DSBS is enabled if } s \le T.
\label{eq:dbs_fixed_schedule}
$
{ Adaptive strategy based on localization agreement.}
We use the matching quality from Sec.~\ref{sec:dsue} to measure localization maturity. Let
\vspace{-2mm}
\begin{equation}
\bar{\sigma}_{\mathrm{match}}^{(s)}
=
\mathbb{E}_{I\in \mathcal{I}^{(s)}_u}
\left[
\mathbb{E}_{(j,k)\in \mathcal{P}(I)}
\left[
\sigma^{\mathrm{match}}_{jk}
\right]
\right].
\vspace{-2mm}
\label{eq:dbs_adaptive_metric}
\end{equation}
If
$
\bar{\sigma}_{\mathrm{match}}^{(s)} < \tau_{\mathrm{loc}},
$
we enable DSBS; otherwise, we disable it. This strategy is more robust across datasets and detectors with different convergence speeds.

\subsection{Active Learning Loop }
\label{sec:al_loop_impl}

Let the initial labeled set be $\mathcal{I}^{(0)}_l$, and the initial unlabeled set be
$
\mathcal{I}^{(0)}_u=\mathcal{I}\setminus \mathcal{I}^{(0)}_l.
$
Let the total annotation budget be $B$ (measured in number of images), and the per-round annotation budget be $n$. Then the maximum number of AL iterations is
$
S=\left\lceil \frac{B}{n}\right\rceil.
$
Our AL loop is as follows.

\textbf{(0) Offline SA-source precomputation (one-time only).}
For all $I\in\mathcal{I}$, we precompute and cache the SA-source:
$
\big(\mathcal{B}_{SA}(I),\mathcal{M}_{SA}(I),\mathcal{S}_{SA}(I)\big)
$. This is a one-time cost, and all later AL rounds directly read from the cache.

\textbf{(1) Initial training (step $s=0$).}
Train the detector on the initial labeled set $\mathcal{I}^{(0)}_l$:
$
\mathrm{OD}^{(0)} \leftarrow \mathrm{Train}(\mathrm{OD};\ \mathcal{I}^{(0)}_l).
$

\textbf{(2) Unlabeled inference and FDSU uncertainty evaluation (step $s\ge 1$).}
For every unlabeled image $I\in\mathcal{I}^{(s)}_u$, run inference with the current detector $\mathrm{OD}^{(s-1)}$ to obtain the OD-source predictions
$\big(\mathcal{B}_{OD}(I), \Sigma_{OD}(I)\big)$.
Then combine them with the cached SA-source $\mathcal{B}_{SA}(I)$ and compute FDSU to obtain image-level uncertainty:
$
U(I)\leftarrow
\mathrm{FDSU}\big(\mathcal{B}_{SA}(I),\mathcal{B}_{OD}(I),\Sigma_{OD}(I)\big),
$
.

\textbf{(3) Candidate set expansion (uncertainty filtering).}
Using a candidate expansion factor $\theta>1$, select the top $n\theta$ most uncertain images from the unlabeled pool to form the candidate set:
$
\mathcal{I}^{(s)}_{\mathrm{cond}}
=
\operatorname{Top}_{n\theta}\Big(\{U(I)\}_{I\in\mathcal{I}^{(s)}_u}\Big).
$

\textbf{(4) Diversity sampling (diversity selection).}
Compute the diversity score $D(I)$ on $\mathcal{I}^{(s)}_{\mathrm{cond}}$, and select the top-$n$ images as the final set for annotation in this round:
$
D(I)\leftarrow
\mathrm{Div}\big(I;\ \mathcal{I}^{(s)}_l,\mathcal{B}_{SA}(I),\mathcal{M}_{SA}(I)\big),
\mathcal{I}^{(s)}_{\mathrm{sel}}
=
\operatorname{Top}_{n}\Big(\{D(I)\}_{I\in\mathcal{I}^{(s)}_{\mathrm{cond}}}\Big).
$
Here, $\mathrm{Div}(\cdot)$ uses DINOv2-based mask-guided object embeddings and includes the UPN fragmentation suppression term.

\textbf{(5) DSBS semi-automated labeling and dataset update.}
For each image $I\in\mathcal{I}^{(s)}_{\mathrm{sel}}$, we use DSBS for pre-labeling: by default, the bounding box comes from the SA-source $\mathcal{B}_{SA}(I)$, while the category comes from detector predictions, and the annotator confirms/corrects them. After annotation, we update the labeled/unlabeled sets:
$
\mathcal{I}^{(s+1)}_l
=
\mathcal{I}^{(s)}_l \cup \mathcal{I}^{(s)}_{\mathrm{sel}},
\mathcal{I}^{(s+1)}_u
=
\mathcal{I}\setminus \mathcal{I}^{(s+1)}_l.
$
Fine-tune the detector on the updated labeled set $\mathcal{I}^{(s+1)}_l$ to obtain $\mathrm{OD}^{(s)}$:
$
\mathrm{OD}^{(s)}
\leftarrow
\mathrm{FineTune}\big(\mathrm{OD}^{(s-1)};\ \mathcal{I}^{(s+1)}_l\big).
$
Repeat steps (2)--(6) until the total budget $B$ is exhausted or the unlabeled set empty.

\section{Experiments}
\vspace{-1mm}
\subsection{Datasets and Settings}
\label{sec:implementation_details}
\textbf{Datasets}
We evaluate the proposed framework on four public object detection datasets, including DIOR \cite{zhan2023rsvg} (with diverse object categories and scene types), DOTAv2 (which contains a large number of extremely small object annotations, $\leq 10$ pixels), FAIR1M \cite{sun2022fair1m} (a fine-grained remote sensing object detection dataset), and HRSC2016 \cite{tang2020h} (a ship detection dataset). DIOR \cite{zhan2023rsvg} and HRSC2016 provide predefined train/validation/test splits. Since the test split of DOTAv2 does not provide public labels, we adopt the data split proposed by Lee et al \cite{lee2022interactive}. For FAIR1M \cite{sun2022fair1m}, we randomly split the dataset into 40\%/20\%/20\% for training, validation, and testing.

\textbf{Model Settings:} We follow the default hyperparameters and training configurations of RTDETRV2 \cite{lv2024rt}, and use a ResNet50 backbone pretrained on FMOW.
In implementation, we adopt a one-time offline caching strategy for the SA-source, where reference boxes and masks are pre-stored for each image. We further control storage and subsequent computation overhead through Top-$K_u$ retention and lightweight geometric filtering, thereby avoiding repeated UPN and SAM2 inference in each active learning round. In the diversity sampling stage, we build mask-guided object embeddings using DINOv2 feature maps and SAM2 masks, and extract/cache features on demand for candidate images. Finally, DSBS is most effective during the cold-start stage, and can be scheduled either with a fixed-round strategy or an adaptive strategy based on global matching quality.

\begin{figure}[t]
\begin{center}
\includegraphics[width=8.5cm, height=6cm]{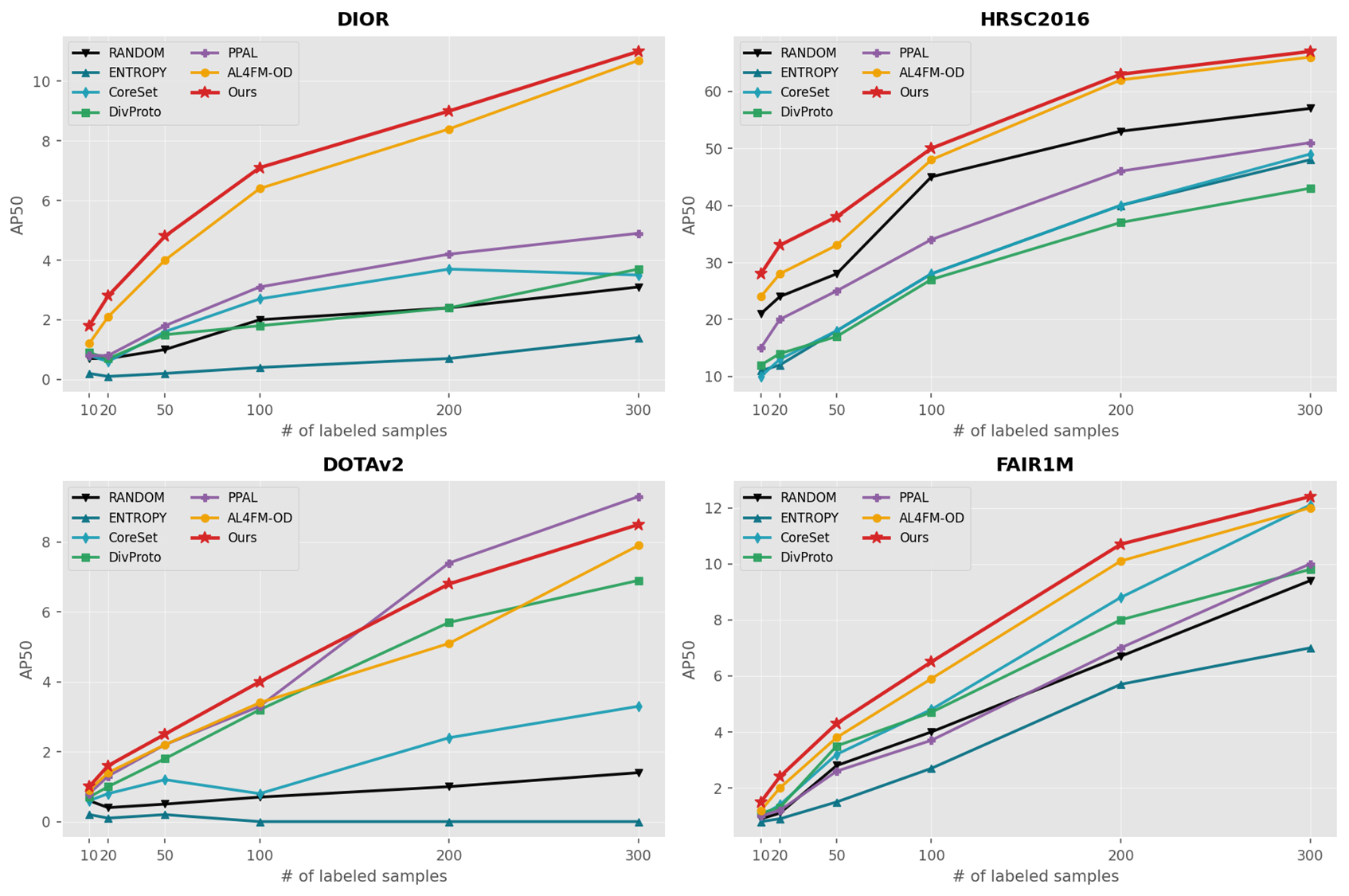}
\end{center}
\vspace{-6 mm}
\caption{Comparison of improvements under different annotation budgets on DIOR  \cite{zhan2023rsvg}, HRSC2016 \cite{tang2020h}, DOTAv2, and FAIR1M \cite{sun2022fair1m}. Entropy \cite{yang2024plug}, CoreSet  \cite{lee2024coreset}, PPAL \cite{yang2024plug}, DivProto \cite{wu2022entropy}, AL4FM \cite{burges2025active}}
\label{result}
\vspace{-7 mm}
\end{figure}

\subsection{Experimental Results}

\textbf{Comparison with state-of-the-art methods:}
As shown in Fig.~\ref{result}, even without DSBS, our method achieves the best or competitive performance under most annotation budgets, with particularly clear advantages in the low-budget regime. This demonstrates that the proposed sampling framework can more reliably identify informative samples during cold start and improve the utilization of limited annotation budgets.
On DIOR~\cite{zhan2023rsvg} and HRSC2016~\cite{tang2020h}, our method consistently performs favorably across different budgets, with larger gains at 10, 20, and 50 labeled images. These results indicate that the localization-stable SA-source, combined with FDSU, reduces the influence of noisy detector boxes on early-stage sample selection. On FAIR1M~\cite{sun2022fair1m}, our method maintains leading performance in the medium- and high-budget regimes, suggesting that DINOv2-based mask-guided object representations improve diversity coverage in complex multi-class scenes. On DOTAv2, our method outperforms AL4FM-OD under low and medium budgets and remains competitive at higher budgets, although it is slightly below PPAL at the largest budget.
This comparison evaluates only the active learning sampling strategies and excludes DSBS, whose primary role is to accelerate semi-automatic annotation and reduce human correction effort. Its impact on cold-start annotation efficiency is evaluated separately in the following section.
\begin{figure}[t]
\begin{center}
\includegraphics[width=6cm, height=3cm]{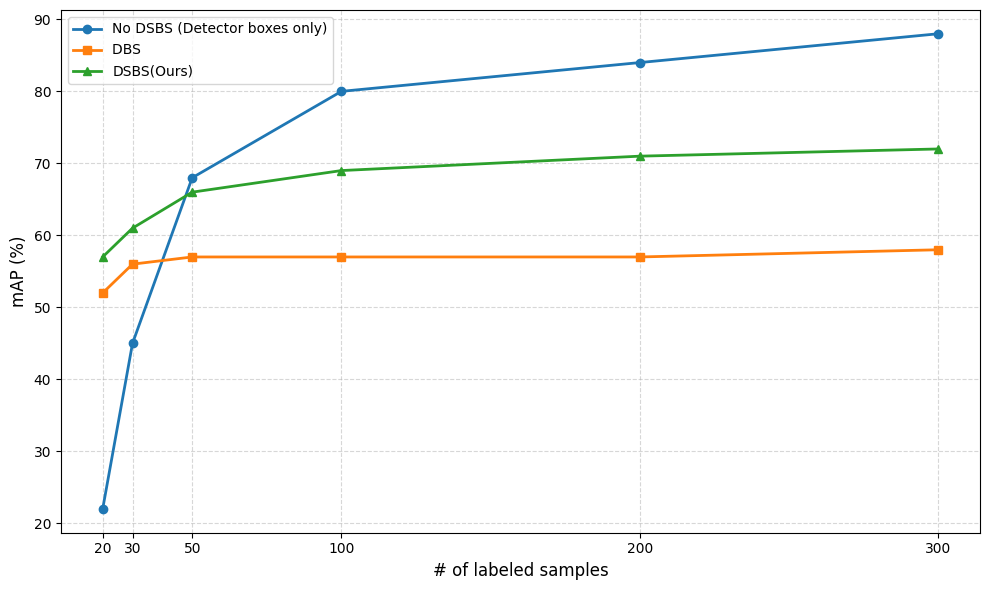}
\end{center}
\vspace{-6 mm}
\caption{Cold-start results under different methods on HRSC2016 \cite{tang2020h}.}
\label{dsbs}
\vspace{-3mm}
\end{figure}
\textbf{Analysis of  DSBS in the Cold-Start Stage: }
We evaluate using {Recall@100} with a stricter {IoU threshold of 75\%}. As shown in Fig \ref{dsbs}, {DSBS (Ours)} substantially improves cold-start performance under low annotation budgets. With only {20/30 annotated samples}, DSBS achieves about {57\%/61\% mAP}, outperforming {No DSBS} ({22\%/45\%}) and {DBS} ({52\%/56\%}). This indicates that, under extremely limited supervision, using the {SA-source} as a stable geometric prior with dual-source box switching effectively mitigates early-stage localization noise and enables faster attainment of usable detection performance.
These results show that DSBS is most beneficial in the low-budget regime, where improved box initialization reduces early training instability. Overall, DSBS serves as a practical semi-automatic annotation module for cold-start active learning by lowering the annotation cost required to reach usable performance and improving early-stage annotation efficiency.
\begin{table}[t]
\centering
\small
\resizebox{0.48\textwidth}{!}{%
\begin{tabular}{llccccc}
\toprule
\multirow{2}{*}{\textbf{Stage 1}} & \multirow{2}{*}{\textbf{Stage 2}} & \multicolumn{5}{c}{\textbf{mAP on \# of labeled images}} \\
\cmidrule(lr){3-7}
 &  & \textbf{10} & \textbf{50} & \textbf{100} & \textbf{200} & \textbf{300} \\
\midrule
\multicolumn{2}{c}{Random}      & $0.3\!\pm\!0.2$ & $1.2\!\pm\!0.6$ & $1.6\!\pm\!0.6$ & $2.3\!\pm\!0.5$ & $4.0\!\pm\!0.7$ \\
\midrule
Entropy & None                  & $0.5\!\pm\!0.3$ & $0.2\!\pm\!0.1$ & $0.4\!\pm\!0.4$ & $0.7\!\pm\!0.4$ & $1.3\!\pm\!1.1$ \\
DSUE    & None                  & $0.9\!\pm\!0.4$ & $2.4\!\pm\!0.9$ & $3.8\!\pm\!0.5$ & $4.3\!\pm\!0.9$ & $5.4\!\pm\!0.8$ \\
FDSU    & None                  & $\mathbf{1.1\!\pm\!0.3}$ & $\mathbf{2.7\!\pm\!0.8}$ & $\mathbf{4.3\!\pm\!0.4}$ & $\mathbf{4.8\!\pm\!0.7}$ & $\mathbf{6.0\!\pm\!0.5}$ \\
\midrule
FDSU    & CoreSet               & $0.9\!\pm\!0.4$ & $2.9\!\pm\!0.7$ & $4.4\!\pm\!1.0$ & $4.8\!\pm\!0.4$ & $5.0\!\pm\!0.6$ \\
FDSU    & DivProto              & $1.0\!\pm\!0.4$ & $1.5\!\pm\!0.6$ & $1.0\!\pm\!0.5$ & $1.3\!\pm\!0.5$ & $1.4\!\pm\!0.5$ \\
FDSU    & CCMS                  & $0.9\!\pm\!0.3$ & $2.0\!\pm\!1.1$ & $4.0\!\pm\!0.5$ & $4.9\!\pm\!0.8$ & $5.6\!\pm\!0.8$ \\
FDSU    & MDE                   & $1.0\!\pm\!0.3$ & $4.4\!\pm\!0.8$ & $6.8\!\pm\!1.3$ & $8.8\!\pm\!1.6$ & $11.0\!\pm\!0.7$ \\
FDSU    & \textbf{OCDS}         & $\mathbf{1.2\!\pm\!0.3}$ & $\mathbf{4.7\!\pm\!0.8}$ & $\mathbf{7.2\!\pm\!1.2}$ & $\mathbf{9.1\!\pm\!1.5}$ & $\mathbf{11.3\!\pm\!0.6}$ \\
\bottomrule
\end{tabular}}
\caption{Ablation studies of FDSU on DIOR \cite{zhan2023rsvg}. 
CCMS is from PPAL; DSUE and MDE used in AL4FM-OD; OCDS is ours.}
\label{tab:stage1_stage2_ablation}
\vspace{-11mm}
\end{table}

\begin{table*}[t]
\centering
\small
\resizebox{1\textwidth}{!}{%
\begin{tabular}{l|cc|cc|c|cc|ccccc}
\toprule
\multirow{2}{*}{\textbf{Dataset}} &
\multicolumn{2}{c|}{\textbf{UPN (Top-50)}} &
\multicolumn{2}{c|}{\textbf{UPN (Top-100)}} &
\multicolumn{1}{c|}{\textbf{UPN (Top-200)}} &
\multicolumn{2}{c|}{\textbf{UPN+SAM2 (Top-200)}} &
\multicolumn{5}{c}{\textbf{Fragmentation Suppression}} \\
\cmidrule(lr){2-3}\cmidrule(lr){4-5}\cmidrule(lr){6-6}\cmidrule(lr){7-8}\cmidrule(lr){9-13}
& \textbf{R@0.5} & \textbf{R@0.7}
& \textbf{R@0.5} & \textbf{R@0.7}
& \textbf{R@0.5 / R@0.7}
& \textbf{R@0.5 / R@0.7} & \textbf{Avg $\Delta$IoU}
& \textbf{Red. (w/o)} & \textbf{Red. (w/)}
& \textbf{Frag. (w/o)} & \textbf{Frag. (w/)}
& \textbf{Reduction (\%)} \\
\midrule
DIOR \cite{zhan2023rsvg}
& 78.6 & 51.2
& 86.9 & 60.8
& 91.7 / 68.4
& 94.8 / 74.9 & 4.7
& 0.41 & 0.27
& 0.36 & 0.21
& 34.1 \\
HRSC2016 \cite{tang2020h}
& 89.8 & 63.5
& 94.6 & 72.9
& 97.1 / 80.3
& 98.3 / 86.7 & 6.5
& 0.29 & 0.18
& 0.22 & 0.12
& 37.9 \\
DOTAv2
& 62.4 & 31.7
& 73.8 & 40.9
& 82.7 / 50.6
& 87.9 / 58.8 & 5.9
& 0.58 & 0.36
& 0.49 & 0.29
& 37.9 \\
FAIR1M \cite{sun2022fair1m}
& 75.1 & 46.8
& 84.5 & 56.4
& 90.2 / 64.9
& 93.6 / 71.8 & 5.1
& 0.46 & 0.30
& 0.39 & 0.24
& 34.8 \\
\bottomrule
\end{tabular}}
\caption{Results of the foundation pipeline quality analysis on 4 different datasets. Red: Redundancy, and Frag: Small-fragment.}
\label{tab:foundation_pipeline_quality_results}
\vspace{-11mm}
\end{table*}

\subsection{Ablations Studies}
The ablation results on the DIOR~\cite{zhan2023rsvg} dataset are shown in Table \ref{tab:stage1_stage2_ablation}. We first compare different Stage-1 uncertainty strategies when no second-stage re-ranking is used. Compared with Entropy and DSUE, the proposed FDSU achieves better results under all annotation budgets, with especially larger gains in the low-budget regime. This indicates that the  FDSU, built upon a stronger SA-source (UPN+SAM2), can evaluate sample value more reliably, thereby improving sample selection quality and sample efficiency in the cold-start stage.
Furthermore, with Stage 1 fixed to FDSU, we compare the effects of different Stage-2 strategies (CoreSet, DivProto, CCMS, MDE, and our OCDS). The results show that weaker diversity strategies (CoreSet and DivProto) provide limited gains, while stronger second-stage strategies (MDE) significantly improve performance in the medium- to high-budget regime. Notably, FDSU + OCDS achieves the best results at all budget points and consistently outperforms FDSU + MDE overall. This demonstrates that the proposed OCDS works synergistically with FDSU: the former improves second-stage selection quality through object-level semantic coverage and fragmentation suppression, while the latter improves first-stage candidate quality through more stable dual-source uncertainty estimation. 

\textbf{Foundation Pipeline Quality Analysis.}
Table~\ref{tab:foundation_pipeline_quality_results} validates the effectiveness of the proposed foundation-model pipeline. UPN achieves strong category-agnostic proposal recall across all four datasets, with both R@0.5 and R@0.7 increasing consistently as the number of retained candidates grows from Top-50 to Top-200. HRSC2016~\cite{tang2020h} achieves the highest recall, suggesting that UPN is particularly effective for ship scenes with relatively consistent object structures. In contrast, DOTAv2 shows lower recall at Top-200 (82.7\%/50.6\%), reflecting the difficulty of handling dense small objects and large orientation variations.
Using UPN proposals as prompts, SAM2 further improves refined-box recall on all datasets, especially at the stricter IoU threshold. This demonstrates that SAM2 enhances proposal localization quality and strengthens the SA-source used by FDSU and DSBS.
The fragmentation suppression strategy also consistently reduces both the Redundancy Ratio and Small-fragment Ratio, yielding an overall reduction of approximately 34\%--38\%. This confirms that it effectively suppresses repeated and fragmented proposals that would otherwise introduce pseudo-diversity during sample selection.
Overall, the collaborative pipeline provides reliable candidates for active learning and semi-automatic annotation.

\vspace{-2mm}
\section{Conclusion}
This paper proposes a foundation-model-assisted active learning and semi-automatic annotation framework for efficient remote sensing object detection dataset construction. Experimental results show that the proposed method achieves performance superior to or comparable with existing methods across multiple remote sensing object detection datasets under most annotation budgets. In addition, we propose DSBS for semi-automatic annotation acceleration in the cold-start stage, which effectively reduces the manual burden of box refinement. Overall, this work validates the effectiveness of integrating vision foundation models into the active learning pipeline and provides a new perspective for improving data construction efficiency in remote sensing object detection. Future work will explore adaptive DSBS switching strategies, stronger candidate de-redundancy mechanisms, and extensions to more complex settings.
\section*{Acknowledgment}
This material is based upon work supported by the Air Force Research Laboratory. Any opinions, findings, conclusions, or recommendations expressed in this publication are those of the authors and do not necessarily reflect the views of the U.S. Air Force. The U.S. Government is authorized to reproduce and distribute reprints for Governmental purposes notwithstanding any copyright notation thereon. Approved for Public Release; Distribution Unlimited: AFR/PA Case No. AFRL-2026-3111. 
\bibliographystyle{ieee_fullname}
\bibliography{egbib}

@inproceedings{caron2021emerging,
  title={Emerging properties in self-supervised vision transformers},
  author={Caron, Mathilde and Touvron, Hugo and Misra, Ishan and J{\'e}gou, Herv{\'e} and Mairal, Julien and Bojanowski, Piotr and Joulin, Armand},
  booktitle={ICCV},
  year={2021}
}

@article{oquab2023dinov2,
  title={Dinov2: Learning robust visual features without supervision},
  author={Oquab, Maxime and Darcet, Timoth{\'e}e and Moutakanni, Th{\'e}o and Vo, Huy and Szafraniec, Marc and Khalidov, Vasil and Fernandez, Pierre and Haziza, Daniel and Massa, Francisco and El-Nouby, Alaaeldin and others},
  journal={arXiv preprint arXiv:2304.07193},
  year={2023}
}

@inproceedings{burges2025active,
  title={Active learning meets foundation models: fast remote sensing data annotation for object detection},
  author={Burges, Marvin and Dias, Philipe Ambrozio and Woody, Carson and Walters, Sarah and Lunga, Dalton},
  booktitle={ICCV},
  year={2025}
}

@inproceedings{lee2024coreset,
  title={Coreset selection for object detection},
  author={Lee, Hojun and Kim, Suyoung and Lee, Junhoo and Yoo, Jaeyoung and Kwak, Nojun},
  booktitle={CVPR},
  year={2024}
}

@inproceedings{wu2022entropy,
  title={Entropy-based active learning for object detection with progressive diversity constraint},
  author={Wu, Jiaxi and Chen, Jiaxin and Huang, Di},
  booktitle={CVPR},
  year={2022}
}

@article{zhan2023rsvg,
  title={Object detection in optical remote sensing images: A survey and a new benchmark},
  author={Li, Ke and Wan, Gang and Cheng, Gong and Meng, Liqiu and Han, Junwei},
  journal={ISPRS journal of photogrammetry and remote sensing},
  volume={159},
  pages={296--307},
  year={2020},
  publisher={Elsevier}
}

@article{bar2024active,
  title={Active learning via classifier impact and greedy selection for interactive image retrieval},
  author={Bar, Leah and Lerner, Boaz and Darshan, Nir and Ben-Ari, Rami},
  journal={arXiv preprint arXiv:2412.02310},
  year={2024}
}

@article{sun2022fair1m,
  title={FAIR1M: A benchmark dataset for fine-grained object recognition in high-resolution remote sensing imagery},
  author={Sun, Xian and Wang, Peijin and Yan, Zhiyuan and Xu, Feng and Wang, Ruiping and Diao, Wenhui and Chen, Jin and Li, Jihao and Feng, Yingchao and Xu, Tao and others},
  journal={ISPRS},
  volume={184},
  year={2022},
  publisher={Elsevier}
}

@inproceedings{yoo2019learning,
  title={Learning loss for active learning},
  author={Yoo, Donggeun and Kweon, In So},
  booktitle={CVPR},
  year={2019}
}

@inproceedings{yuan2021multiple,
  title={Multiple instance active learning for object detection},
  author={Yuan, Tianning and Wan, Fang and Fu, Mengying and Liu, Jianzhuang and Xu, Songcen and Ji, Xiangyang and Ye, Qixiang},
  booktitle={CVPR},
  year={2021}
}

@article{tang2020h,
  title={H-YOLO: A single-shot ship detection approach based on region of interest preselected network},
  author={Tang, Gang and Liu, Shibo and Fujino, Iwao and Claramunt, Christophe and Wang, Yide and Men, Shaoyang},
  journal={Remote Sensing},
  volume={12},
  number={24},
  pages={4192},
  year={2020},
  publisher={MDPI}
}

@article{jing2024object,
  title={Object Recognition Consistency in Regression for Active Detection},
  author={Jing, Ming and Ou, Zhilong and Wang, Hongxing and Li, Jiaxin and Zhao, Ziyi},
  journal={Machine Vision and Applications},
  volume={35},
  number={5},
  pages={121},
  year={2024},
  publisher={Springer}
}

@inproceedings{lee2022interactive,
  title={Interactive multi-class tiny-object detection},
  author={Lee, Chunggi and Park, Seonwook and Song, Heon and Ryu, Jeongun and Kim, Sanghoon and Kim, Haejoon and Pereira, S{\'e}rgio and Yoo, Donggeun},
  booktitle={CVPR},
  pages={14136--14145},
  year={2022}
}

@article{sener2017active,
  title={Active learning for convolutional neural networks: A core-set approach},
  author={Sener, Ozan and Savarese, Silvio},
  journal={arXiv preprint},
  year={2017}
}

@inproceedings{yang2024plug,
  title={Plug and play active learning for object detection},
  author={Yang, Chenhongyi and Huang, Lichao and Crowley, Elliot J},
  booktitle={CVPR},
  year={2024}
}

@article{lv2024rt,
  title={Rt-detrv2: Improved baseline with bag-of-freebies for real-time detection transformer},
  author={Lv, Wenyu and Zhao, Yian and Chang, Qinyao and Huang, Kui and Wang, Guanzhong and Liu, Yi},
  journal={arXiv preprint arXiv:2407.17140},
  year={2024}
}

@article{wang2017incorporating,
  title={Incorporating diversity and informativeness in multiple-instance active learning},
  author={Wang, Ran and Wang, Xi-Zhao and Kwong, Sam and Xu, Chen},
  journal={IEEE transactions on fuzzy systems},
  volume={25},
  number={6},
  pages={1460--1475},
  year={2017},
  publisher={IEEE}
}

@article{yang2015multi,
  title={Multi-class active learning by uncertainty sampling with diversity maximization},
  author={Yang, Yi and Ma, Zhigang and Nie, Feiping and Chang, Xiaojun and Hauptmann, Alexander G},
  journal={IJCV},
  volume={113},
  number={2},
  pages={113--127},
  year={2015},
  publisher={Springer}
}

@inproceedings{agarwal2020contextual,
  title={Contextual diversity for active learning},
  author={Agarwal, Sharat and Arora, Himanshu and Anand, Saket and Arora, Chetan},
  booktitle={ECCV},
  year={2020},
  organization={Springer}
}

@inproceedings{wang2023alwod,
  title={ALWOD: Active learning for weakly-supervised object detection},
  author={Wang, Yuting and Ilic, Velibor and Li, Jiatong and Kisa{\v{c}}anin, Branislav and Pavlovic, Vladimir},
  booktitle={ICCV},
  year={2023}
}

@article{chen2022making,
  title={Making your first choice: To address cold start problem in vision active learning},
  author={Chen, Liangyu and Bai, Yutong and Huang, Siyu and Lu, Yongyi and Wen, Bihan and Yuille, Alan L and Zhou, Zongwei},
  journal={arXiv preprint arXiv:2210.02442},
  year={2022}
}

@article{simeoni2025dinov3,
  title={Dinov3},
  author={Sim{\'e}oni, Oriane and Vo, Huy V and Seitzer, Maximilian and Baldassarre, Federico and Oquab, Maxime and Jose, Cijo and Khalidov, Vasil and Szafraniec, Marc and Yi, Seungeun and Ramamonjisoa, Micha{\"e}l and others},
  journal={arXiv preprint arXiv:2508.10104},
  year={2025}
}

@inproceedings{kirillov2023segment,
  title={Segment anything},
  author={Kirillov, Alexander and Mintun, Eric and Ravi, Nikhila and Mao, Hanzi and Rolland, Chloe and Gustafson, Laura and Xiao, Tete and Whitehead, Spencer and Berg, Alexander C and Lo, Wan-Yen and others},
  booktitle={ICCV},
  year={2023}
}

@article{ravi2024sam,
  title={Sam 2: Segment anything in images and videos},
  author={Ravi, Nikhila and Gabeur, Valentin and Hu, Yuan-Ting and Hu, Ronghang and Ryali, Chaitanya and Ma, Tengyu and Khedr, Haitham and R{\"a}dle, Roman and Rolland, Chloe and Gustafson, Laura and others},
  journal={arXiv preprint arXiv:2408.00714},
  year={2024}
}

@article{jiang2024chatrex,
  title={Chatrex: Taming multimodal llm for joint perception and understanding},
  author={Jiang, Qing and Luo, Gen and Yang, Yuqin and Xiong, Yuda and Chen, Yihao and Zeng, Zhaoyang and Ren, Tianhe and Zhang, Lei},
  journal={arXiv preprint arXiv:2411.18363},
  year={2024}
}

@inproceedings{carion2020end,
  title={End-to-end object detection with transformers},
  author={Carion, Nicolas and Massa, Francisco and Synnaeve, Gabriel and Usunier, Nicolas and Kirillov, Alexander and Zagoruyko, Sergey},
  booktitle={ECCV},
  pages={213--229},
  year={2020},
  organization={Springer}
}

@inproceedings{jiang2024t,
  title={T-rex2: Towards generic object detection via text-visual prompt synergy},
  author={Jiang, Qing and Li, Feng and Zeng, Zhaoyang and Ren, Tianhe and Liu, Shilong and Zhang, Lei},
  booktitle={ECCV},
  pages={38--57},
  year={2024},
  organization={Springer}
}

@inproceedings{xiaopath,
  title={Path Matters: Unveiling Geometric Implicit Bias via Curvature-Aware Sparse View Optimization},
  author={Xiao, Canran and Fan, Liaoyuan and Li, Yanbin and Tang, Jing and Yu, Peilai},
  booktitle={ICLR},
 year={2026}
}

@inproceedings{xiao2026reversible,
  title={Reversible primitive--composition alignment for continual vision--language learning},
  author={Xiao, Canran and Xu, Tianxiang and Ma, Siyuan and Jiang, Yiyang and Gao, Haoyu and Wu, Yuhan},
  booktitle={ICLR},
  year={2026}
}

@inproceedings{xiao2026prototype,
  title={Prototype-Aligned Federated Soft-Prompts for Continual Web Personalization},
  author={Xiao, Canran and Hou, Liwei},
  booktitle={WWW},
  year={2026}
}

@inproceedings{xiao2026points,
  title={From points to coalitions: Hierarchical contrastive shapley values for prioritizing data samples},
  author={Xiao, Canran and Dou, Jiabao and Lin, Zhiming and Ke, Zong and Hou, Liwei},
  booktitle={AAAI},
  year={2026}
}

@article{li2024comae,
  title={COMAE: COMprehensive Attribute Exploration for Zero-shot Hashing},
  author = {Li, Yuqi and Long, Qingqing and Zhou, Yihang and Zhang, Ran and Ning, Zhiyuan and Zhu, Zhihong and Zhou, Yuanchun and Wang, Xuezhi and Xiao, Meng},
  journal={ICMR},
  year={2025}
}

@misc{tian2026curvatureadaptiveconsistencyflowmatching,
      title={Curvature-Adaptive Consistency Flow Matching: Autonomous Trajectory Optimization via Reinforcement Learning}, 
      author={Songtao Tian and Guhan Chen and Bohan Li and Jingyi Ma and Zixiong Yu},
      year={2026},
      eprint={2606.22394},
      archivePrefix={arXiv},
      primaryClass={cs.CV},
      url={https://arxiv.org/abs/2606.22394}, 
}

@article{zhang2025adaptive,
  title={Adaptive Event Stream Slicing for Open-Vocabulary Event-Based Object Detection via Vision-Language Knowledge Distillation},
  author={Zhang, Jinchang and Li, Zijun and Lin, Jiakai and Lu, Guoyu},
  journal={arXiv preprint arXiv:2510.00681},
  year={2025}
}

@inproceedings{lin2025keypoint,
  title={Keypoint detection and description for raw bayer images},
  author={Lin, Jiakai and Zhang, Jinchang and Lu, Guoyu},
  booktitle={2025 IEEE International Conference on Robotics and Automation (ICRA)},
  pages={11736--11742},
  year={2025},
  organization={IEEE}
}

@article{fu2026dav,
  title={Dav-gswt: Diffusion-active-view sampling for data-efficient gaussian splatting wang tiles},
  author={Fu, Rong and Wu, Jiekai and Wei, Haiyun and Jia, Yee Tan and Li, Yang and Ma, Xiaowen and Wu, Wangyu and Fong, Simon},
  journal={arXiv preprint arXiv:2602.15355},
  year={2026}
}

@inproceedings{zhang2025vision,
  title={Vision-language embodiment for monocular depth estimation},
  author={Zhang, Jinchang and Lu, Guoyu},
  booktitle={Proceedings of the Computer Vision and Pattern Recognition Conference},
  pages={29479--29489},
  year={2025}
}

@inproceedings{lin20253d,
  title={3D Plant Root Skeleton Detection and Extraction},
  author={Lin, Jiakai and Zhang, Jinchang and Jin, Ge and Song, Wenzhan and Liu, Tianming and Lu, Guoyu},
  booktitle={2025 IEEE/RSJ International Conference on Intelligent Robots and Systems (IROS)},
  pages={3011--3017},
  year={2025},
  organization={IEEE}
}

\end{document}